\documentclass[letterpaper]{article}
\usepackage{natbib,alifeconf,amsmath, hyperref}

\title{Neural Cellular Automata and Deep Equilibrium Models}
\author{Zhibai Jia\\
\mbox{}\\
Cornell University\\
zj263@cornell.edu}

\begin{document}
\maketitle

\begin{abstract}
This essay discusses the connections and differences between two emerging paradigms in deep learning, namely Neural Cellular Automata and Deep Equilibrium Models, and train a simple Deep Equilibrium Convolutional model to demonstrate the inherent similarity of NCA and DEQ based methods. Finally, this essay speculates about ways to combine theoretical and practical aspects of both approaches for future research. 
\end{abstract}

\section{Neural Cellular Automata}

Neural Cellular Automata (NCA) \citep{mordvintsevGrowingNeuralCellular2020} is an emerging deep learning paradigm that seeks to harness self-organization to extend the neural regime. Typically, an NCA is a cellular automaton that has a embedding-style latent state, and a state transition rule parameterized by a neural network, trained by end-to-end backpropagation. The theory of NCAs is generally founded on the theory of complex systems, and draws heavily from the literature of classical cellular automata, pioneered and popularized in past decades by Conway\citep{games1970fantastic}, Wolfram \citep{Wolfram2002}, etc, to introduce desirable aspects of cellular modeling, such as parameter efficiency, structure flexibility, and emergent behaviour into modern deep learning.\\ 
Notably, the Google Team of Mordvintsev et al. have successfully applied NCA to morphogenic modelling \citep{mordvintsevGrowingNeuralCellular2020}, localized MNIST classification \citep{SelfclassifyingMNISTDigits}, and compact texture generation\citep{mordvintsevMuNCATexture2021}, as well as developing a theoretical framework for building isotropic NCAs \citep{mordvintsevGrowingIsotropicNeural2022}. They host a "Self-Organizing Systems Thread" dedicated to publications on related topics\citep{SelfOrganisingSystems}. Others lines of research have investigated NCA for image segmentation \citep{sandlerImageSegmentationCellular2020} and generation \citep{tesfaldetAttentionbasedNeuralCellular2022}, as well as conjoining NCAs with other well-established deep learning paradigms, such as variational autoencoding\citep{palmVariationalNeuralCellular2022}, graph neural networks\citep{grattarolaLearningGraphCellular2021}, and Vision Transformers\citep{tesfaldetAttentionbasedNeuralCellular2022}. A kind of Meta-Algorithm for learning a functional prior over the space of cellular automata is also described under the name of Neural Cellular Automata Manifold \citep{ruizNeuralCellularAutomata2021}. \\
Generally, these preliminary investigations on Neural Cellular Automata may seem theoretical, or even imaginative and artistic sometimes, in the way that research on emergence often do. (Conway's Game of Life and Wolfram's heavily Visual Dissertation are examples) Besides, it often takes inspiration and motivation from the various disciplines and topics associated with complex systems, instead of orientating towards large scale application in DL. For instance, the seminal work on morphogenetic modeling \citep{mordvintsevGrowingNeuralCellular2020} by Mordvintsev et al. is motivated by biological morphogenesis, yet does not seem to have any direct practical implications. \\
\section{Deep Equilibrium Models}
By contrast, the line of work on deep equilibrium models carry strong engineering and clinical overtones. DEQs are a promising class of implicit models that aim to alleviate the memory bottleneck in deep learning by defining a forward pass to be root-finding process of some parametrized equilibrium function \citep{johnsonChapterDeepEquilibrium}, and update the parameters by approximating an implicit gradient of that process \citep{gengTrainingImplicitModels2022}. Pioneered by Bai et al., DEQs have shown state-of-the-art performance across a spectrum of core DL tasks, such as language modeling\citep{baiDeepEquilibriumModels2019}, image classification\citep{baiMultiscaleDeepEquilibrium2020}, graph modeling\citep{guImplicitGraphNeural2020}, neural ODEs\citep{palContinuousDeepEquilibrium2023}, and implicit neural representations\citep{huangTextbackslashTextrmLbrace2021}, demonstrating memory savings, and perhaps not coincidentally, parameter efficiency. Indeed, despite greatly divergent motivations, research context and community vibes, there are certain deep connections between the two approaches, and I believe it would be beneficial to describe this connection explicitly. \\
\section{Connections Between NCA and DEQ}
The similarity between cellular automata and the CNN convolution have been noted nearly from the dawn of the latter approach in deep learning. In every step, Classical 2D CAs apply a spatially invariant, locally defined transition to every cell or "pixel" in the cell board, naturally evoking a mechanism like a weight-tied convolutional network applied repeatedly through time.\citep{mordvintsevGrowingNeuralCellular2020}\\ While such a architecture is uncommon in the explicit domain, DEQ models exactly utilize this kind of dynamic as its backbone.\citep{johnsonChapterDeepEquilibrium} Specifically, DEQ models that utilize some kind of spatial structure, especially a local operation, such as convolution or graph convolution in their equilibrium function, can be seen as implicitly obtaining the limiting state for a corresponding neural cellular automaton or graph neural cellular automaton. Concurrently, training such a deq model can be viewed as training an NCA to converge to a desired state after an indefinite number of time steps. \\
Gilpin investigated the approximation of a classical CA through such a network, very similar to the inner problem in DEQ models. \citep{gilpinCellularAutomataConvolutional2019} In the classical CA literature, the equilibrium state obtained from a CA after running for a a long time is dubbed "Ash" (Though note that ash is often littered with multiphasal oscillators and non-stationary components, instead of converging statically in the way guaranteed by DEQ theory), hence spatial DEQs could perhaps be alternatively characterized as "Neural Cellular Ash Networks". \\
In DEQs, there is a critical design detail known as input injection, that forces the implicit recurrent dynamics to depend on the input. It is often done residual-style, by directly adding the input to the implicit state. An analogous mechanism exists in NCAs, and is seen in self-classifying MNIST \citep{randazzoSelfclassifyingMNISTDigits2020} and NCA image segmentation \citep{sandlerImageSegmentationCellular2020}, though in this case the common practice is to superimpose the input image as extra channels in the cellular automaton, and is often kept read-only in the dynamic. Of course, when the underlying principle is stated as such, such design differences are in fact trivial, and may seem like mere convention. It will be easy to implement a DEQ with "input superposition" instead of "input injection", or a NCA that has "additive conditioning perturbations"\footnote{I coined this term just now. It is common for the artifical life community to have alternative nomenclature for deep learning devices. Conversely, Mordvintsev et al. noted that the Growing NCA can be characterized in DL jargon as "recurrent convolutions with per-cell dropout"} in its hidden states. In this aspect they are essentially the same.
\\
Finally, if we take the view that cellular automata are in fact a modeling approach for approximating partial differential equations discretely, then NCAs are similar to Neural ODEs in that they learn a complex continuous dynamic from data with a neural network. In this perspective, many spatial DEQs are also solving for a PDE, or integrating them towards infinity. 
\section{Experimental Observation}
We train a very simple implicit CNN to classify MNIST. This model has one explicit conv layer, one convolutional DEQ layer, one global average pooling layer, and finally a two-layer MLP "decoder" to project the pooled hidden activations to labels. \\
\begin{gather}
y = MLP(AvgPool(z^*))\\ 
z^* = z^* + tanh(K_2 * Concat(ReLU(K_1 * x), z^*))
\end{gather}
We implement this model in PyTorch using the TorchDeq \citep{gengTorchDEQLibraryDeep2023} package, and train this model with SGD, with learning rate of 0.005 and momentum set at 0.9. In both training and inference, the hidden recurrent state z is initialized with gaussian random noise. Despite having only 1.5e4 parameters and being trained for 10 epoches, it achieves 97\% accuracy on the test set. Of course, the point of this experiment is not to benchmark any models, but to demonstrate the inherent similarity of DEQ and NCA-based methods. \\
In training, the Broyden method is used to find the root of the equilibrium function quickly. But in our observations, we reprogram the trained model to repeatedly apply the implicit convolutional layer in (2), as in forward equilibrium finding. We visualize after every step the equilibrium condition is applied. 

\begin{figure}[!htb]
\begin{center}
\includegraphics[width=3in]{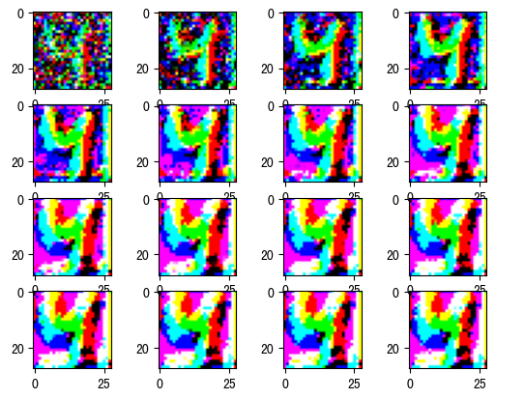}
\caption{Visualizing the hidden channels of the inner iteration in a Deep Equilibrium MNIST classification model}
\label{fig1}
\end{center}
\end{figure}

Unsurprisingly, we observe NCA-like dynamics, where a digit-like pattern emerges from the noise guided by the superposed input injection, and converges to a state where different feature regions are delineated by different stasis combinations of the hidden channels, indicated by different colors in the . This kind of self-organization is also discovered in the hidden state analysis for self-organizing textures.\\ \cite{niklasson2021self-organising}
A interesting property of this model is that it is agnostic to input size, by virtue of the global pooling layer, which cannot be said for most "orthodox" CNNs. If we truncate the input image to only a corner of the original, it can still output a estimate for its label. Conversly, one can imagine a image classification network trained on details naturally generalize to a whole picture or panorama. To obtain the label for one object in the scene, one only needs to sample the steady state at that locality and decode its label. (Somewhat like semantic segmentation)\\

\begin{figure}[!htb]
\begin{center}
\includegraphics[width=3in]{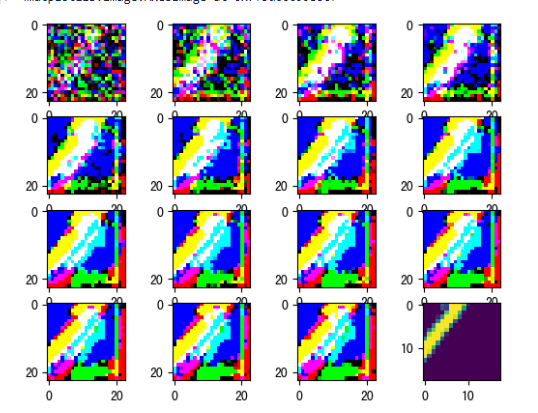}
\caption{A truncated image can still be used to generate the equilibrium state for classification}
\label{fig2}
\end{center}
\end{figure}

\section{Best of Both Worlds}
It is likely that theory and practice from the field of DEQ and NCAs can complement each other. In the NCA regime, Backpropagation Through Time is the dominant way of training, as it is perhaps implied by the "time" analogy in CA literature. Though, BPTT is notoriously memory-consuming and unstable, as well as being "biologically implausible" \footnote{Judging from Mordvintsev et al.'s other interests, such as "Self-Replicating Neural Agents", I would expect them to value biological plausibility}. In fact, in "Growing Neural Cellular Automata", a painstaking three-stage procedure, and the use of a sample pool is required to train the explicit NCA to converge. DEQ-style Implicit differentiation could make building attractors in the NCA space much more efficient. The theory for stability of implicit models developed in the DEQ community is also suitable for avoiding numerical blowups in NCA training. \citep{gengTrainingImplicitModels2022}
On the other hand, theoretical tools from complexity science and cellular automata, such as self-organization and emergence, can be used to characterize the behaviour of spatial DEQs, and would perhaps be useful in explainability investigations. Here I would like to point out that the favoured "path independence" properties of spatial DEQs, including the ability to "exploit extra test-time at inference for more difficult tasks", "training on 9x9 and generalizing to 800x800" \citep{Anil2022PathIE} are all hallmarks shared with NCA methods, and are inherently related to self-organization.\\
The flexibility of NCA can also inspire implicit models in new directions. Most interestingly, it is possible for a model trained as DEQ (root finding used for equilibrium solving) to be deployed as a cellular automata based model (forward computation used), to make use of specialized hardware or even "mortal hardware".



\footnotesize
\bibliographystyle{apalike}
\bibliography{ml}

\end{document}